\documentclass[journal]{IEEEtran}
\usepackage{cite}
\usepackage{booktabs} 
\usepackage[hidelinks]{hyperref}
\usepackage{cite}
\usepackage{textcomp}
\usepackage{wrapfig}
\usepackage{float}
\usepackage{siunitx}
\usepackage{xcolor}
\usepackage{soul}

\usepackage{algorithm}
\usepackage{algpseudocode}
\usepackage{caption}
\usepackage{amsmath,amssymb,amsfonts}
\usepackage{graphicx}
\usepackage{textcomp}
\usepackage{wrapfig}
\usepackage{float}
\usepackage{siunitx}
\usepackage{xcolor}
\definecolor{newcolor}{rgb}{.8,.349,.1}
\usepackage[draft]{changes}
\setdeletedmarkup{\color{red}{\sout{#1}}}
\setaddedmarkup{\color{blue}{#1}}

\def\BibTeX{{\rm B\kern-.05em{\sc i\kern-.025em b}\kern-.08em
    T\kern-.1667em\lower.7ex\hbox{E}\kern-.125emX}}
\definecolor{abstractbg}{rgb}{0.89804,0.94510,0.83137}
\begin{document}
\title{DistillH-Mamba: A Hypergraph-Mamba-Based Knowledge Distillation Model for Efficient Impact Fall Detection}
\author{Tresor Y. Koffi, Youssef Mourchid, Mohammed Hindawi, and Yohan Dupuis
\thanks{Submitted for review on May 15, 2025. This work was supported by CESI, CESI LINEACT, France.}
% \thanks{T. Y. Koffi is a Doctoral at CESI, CESI LINEACT, Dijon, France (e-mail: ytkoffi@cesi.fr).}
\thanks{T. Y. Koffi is a Doctoral Researcher at CESI, CESI LINEACT, Dijon, France, and enrolled at Doctoral School 432 ENSAM ParisTech (e-mail: ytkoffi@cesi.fr).}
\thanks{Y. Mourchid is a Lecturer Researcher at CESI, CESI LINEACT, Dijon, France (e-mail: ymourchid@cesi.fr).}
\thanks{M. Hindawi is a Lecturer Researcher at CESI, CESI LINEACT, Lyon, France (e-mail: mhindawi@cesi.fr).}
\thanks{Y. Dupuis is a Research Director at CESI, CESI LINEACT, Paris, France (e-mail: ydupuis@cesi.fr).}
}

\IEEEtitleabstractindextext{%
\begin{abstract}
Falls among the elderly represent a significant public health concern due to their prevalence, consequences, and societal burden. While deep learning has improved fall detection, accurately identifying impact moments (when an individual hits the ground) remains challenging. Additionally, current algorithms often rely on complex models with high computational demands, limiting real-time deployment feasibility. In this work, we propose DistillH-Mamba, a novel architecture for impact fall detection that addresses these challenges through three key innovations: First, we introduce a hypergraph-based approach that captures higher-order relationships between multiple joints simultaneously, enabling more accurate modeling of complex interactions during impact falls. Second, we integrate the Mamba architecture with hypergraphs for impact detection, significantly accelerating processing speed while efficiently capturing both long-term dependencies and sudden skeletal motion changes. Third, we employ relational knowledge distillation that preserves crucial spatial-temporal relationships while reducing computational demands for real-time impact fall detection.
Evaluated on the 3D Skeletons UP-Fall and UMAFall datasets, our DistillH-Mamba model achieves 97.38\% accuracy in detecting impact within fall events and 73.8\% reduction in inference time compared to its teacher model, outperforming state-of-the-art methods in both precision and efficiency.
\end{abstract}
}

\maketitle
\maketitle

\IEEEdisplaynontitleabstractindextext
\IEEEpeerreviewmaketitle
\section{Introduction}
\label{sec:introduction}

Falls represent a significant public health concern across all age groups, ranking as the second leading cause of unintentional injury-related deaths worldwide \cite{kannus1999fall}. While falls affect individuals of all ages, they pose particularly severe risks for elderly populations, with the global population aged 65 and over projected to grow from 727 million in 2020 to approximately 1.5 billion by 2050 \cite{kannus1999fall}. Developing robust fall detection systems is therefore critical for preventing complications and ensuring timely medical intervention across diverse populations. Immediate assistance following falls can significantly reduce mortality and hospitalization duration, underscoring the need for accurate and reliable detection systems that can generalize across different demographic groups and environmental conditions.

\begin{figure}
\centering  % <- Only centers the image, not the caption
\includegraphics[scale=0.45]{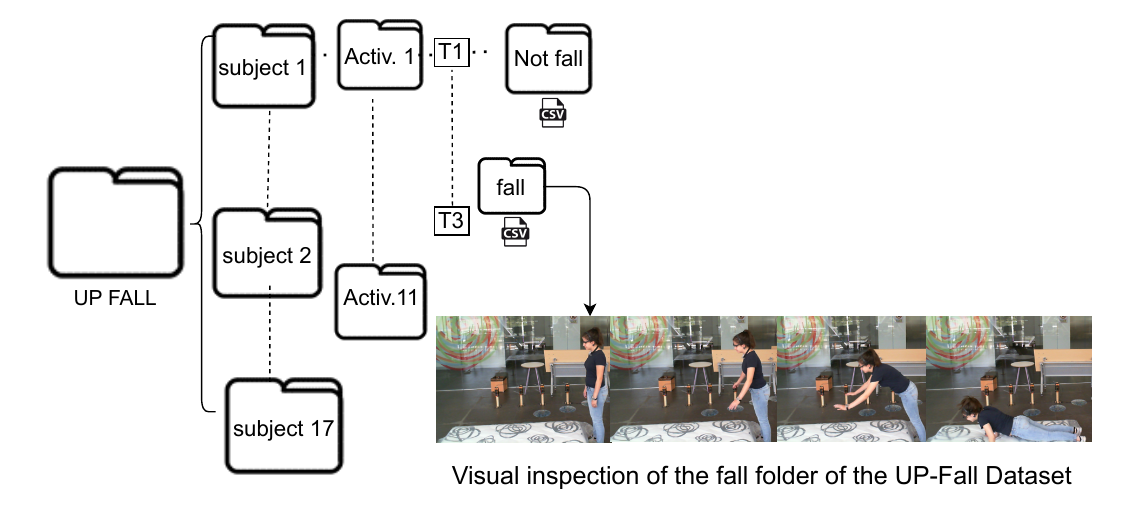}
\caption{Incoherent labelling in UP-Fall Dataset for
 Fall Detection \cite{martinez2019up}}
\label{incoh}
\end{figure}

Various fall detection methods have been explored, broadly divided into traditional and deep learning-based approaches \cite{wang2024fusion}. Traditional threshold-based methods, while straightforward, frequently produce false alarms when encountering movements resembling falls, such as bending or stumbling. Machine learning techniques using multimodal data offer non-intrusive solutions \cite{ren2019research}, but their performance is often limited by poor generalization across diverse scenarios \cite{kottari2019real}.

Deep learning approaches, on the other hand, leverage advanced neural network architectures to process sensor data or images, including 3D information from depth-sensing devices \cite{chen2020fall}. Graph-based deep learning approaches have gained popularity in computer vision for skeleton-based action recognition, as human skeletons naturally align with graph representations\cite{mourchid2016image}. However, traditional Graph Neural Networks (GNNs) are restricted to pairwise joint interactions \cite{zhou2020graph}, failing to capture the complex multi-joint dynamics essential during falls. Hypergraphs overcome this limitation by allowing edges to connect multiple vertices simultaneously, enabling the modeling of higher-order relationships in human motion \cite{dewar2018connectivity}. 
Despite these advancements, accurately detecting the precise moment of impact when an individual hits the ground during a fall remains challenging. Detecting the exact moment of impact during a fall is challenging due to rapid, non-linear changes across multiple joints. These motion patterns can closely resemble non-fall activities such as sitting or sliding, leading to high false-positive rates. Traditional methods often rely on pairwise joint analysis, such as modeling shoulder-elbow or hip-knee relationships independently, which limits the ability to capture the full-body coordination that occurs during a fall. However, accurate impact detection requires recognizing how the moment of ground contact generates a distinct, coordinated pattern across multiple joints. To overcome this limitation, we introduce a hypergraph-based representation that captures higher-order anatomical and functional relationships among multiple joints. Unlike pairwise models, hypergraphs can represent complex interactions by connecting more than two joints within a single hyperedge. This richer structural modeling allows for more accurate differentiation between true falls and similar non-fall movements like controlled sitting or sliding. Furthermore, inconsistencies in dataset labeling, such as in the UP-Fall dataset where fall folders sometimes contain non-fall activities (see Fig. \ref{incoh}), introduce ambiguity that can mislead detection models and reduce accuracy.
Transformer-based models have become popular for action recognition due to their strength in modeling spatial-temporal sequences \cite{nunez2024transformer}. However, their self-attention mechanism introduces quadratic computational complexity, making them resource-intensive and challenging to deploy in real time on constrained devices. Moreover, their global attention distribution often overlooks abrupt, localized transitions such as the critical moment of impact reducing their effectiveness in precise fall detection tasks \cite{zeghina2024deep,pereira2024review}. To address these issues, we adapt the Mamba architecture, which uses selective state spaces to efficiently process long skeletal sequences with linear computational complexity \cite{qu2024survey}. This is especially important for fall detection, where maintaining awareness of a subject's full motion trajectory, from standing posture to ground contact, is critical. Integrating Mamba with our hypergraph-based representation combines efficient temporal modeling with accurate multi-joint spatial reasoning. This design significantly improves the system's ability to distinguish real impacts from non-fall activities, while maintaining computational efficiency. Nevertheless, further optimization is still required to achieve real-time performance on low-power devices.
Knowledge distillation, where a smaller student model learns from a larger teacher model, offers a solution to this efficiency challenge. Traditional distillation methods focus on transferring class probabilities or feature representations \cite{moslemi2024survey}, but preserving temporal and spatial relationships in skeletal data is vital for accurate impact detection \cite{ma2022learning,gunasekara2023joint}. Relational knowledge distillation (RKD) has shown promise in maintaining structural information during model compression \cite{park2019relational}, yet its application to fall detection is unexplored. To address these challenges, we propose DistillH-Mamba, a novel framework that integrates hypergraph representations with Mamba state-space models, enhanced by relational knowledge distillation for computational efficiency. This approach effectively captures higher-order joint relationships and temporal dynamics while remaining lightweight for real-time applications.

The main contributions of this paper are as follows:
\begin{itemize}
   
    \item We propose a dual-representation hypergraph approach that captures higher-order anatomical and functional relationships among multiple joints, enabling precise distinction between actual ground impacts and similar non-fall activities like sitting or sliding.

    \item We introduce the first adoption of Mamba architecture with hypergraphs for impact detection, enabling efficient modeling of temporal dependencies to precisely identify ground contact moments while maintaining linear computational complexity.
    
    \item We develop a relational knowledge distillation technique with novel spatial and temporal loss functions designed for impact detection during falls, preserving critical hypergraph-based joint interactions while creating a lightweight model suitable for resource-constrained devices.

    \item We validate our approach on two datasets, achieving 97.38\% accuracy in impact detection while reducing inference time by 73.8\%, demonstrating both effectiveness and computational efficiency for real-world applications.
\end{itemize}

 \section{Related Works}\label{sec:related_work}

Fall detection methodologies can be broadly categorized into traditional methods and deep learning-based techniques. While these approaches have shown promising results, accurately identifying the impact moment during falls remains challenging, and many existing models require substantial computational resources, limiting real-time deployment capabilities.

\subsection{Traditional Fall Detection Approaches}

Traditional fall detection includes threshold-based techniques and machine learning-based approaches. Threshold-based methods analyze sensors data. These thresholds are empirically determined through prior studies \cite{nooruddin2022sensor}, but vary significantly across individuals due to differences in body characteristics and movement patterns, reducing accuracy and adaptability \cite{vallabh2016fall}. A critical limitation is the difficulty in distinguishing the precise impact moment, which is essential for timely intervention. In contrast,machine learning (ML)-based approaches analyze sensor or image data to identify patterns and classify fall events. Various algorithms including KNN, SVM, DT, and GBOOST have been applied to fall detection \cite{wang2024fusion}, with KNN demonstrating superior performance. In our previous work \cite{koffi2023machine}, we explored the use of ML techniques to detect impact moments within fall events, demonstrating feasibility for addressing this challenge. However, this initial study relied solely on accelerometer data, requiring subjects to continuously wear sensors, which can be intrusive and impractical for everyday use, particularly among elderly individuals.

\subsection{Deep Learning Approaches for Fall Detection}

Deep learning has emerged as a powerful tool for fall detection, offering significant improvements over traditional methods by automatically learning complex patterns from data \cite{chhetri2021deep}. These approaches can process raw sensor data or images and extract meaningful representations, making them well-suited for modeling the complex dynamics of human motion during falls \cite{espinosa2019vision}.
Various neural network architectures have been applied to fall detection, each with distinct advantages. Convolutional Neural Networks (CNNs) excel at capturing spatial features from sensor data or images, while Recurrent Neural Networks (RNNs), particularly Long Short-Term Memory (LSTM) networks, effectively model temporal dependencies in sequential data. Combined CNN-LSTM architectures have demonstrated enhanced performance by simultaneously modeling both spatial and temporal aspects of falls \cite{wu2021fall}.
Spatial-Temporal Graph Convolutional Networks (STGCNs) represent an important advancement by explicitly modeling relationships between skeletal joints \cite{liang2024skeleton}. While effective for skeleton-based fall detection, STGCNs are fundamentally limited by their reliance on pairwise relationships between joints, as traditional graphs only model connections between two nodes at a time \cite{wu2021fall}. This restriction hinders their ability to capture higher-order interactions among multiple joints crucial for accurately modeling complex fall dynamics, particularly during the critical impact phase.

\subsection{Knowledge Distillation for Efficient Model}

As deep learning models for fall detection grow in complexity to achieve higher accuracy, their computational and memory demands increase significantly, challenging real-time deployment on resource-constrained devices. Knowledge distillation (KD) addresses this challenge by transferring knowledge from a complex model (teacher) to a smaller model (student) while maintaining performance \cite{moslemi2024survey}.
Several KD approaches have been applied to fall detection. Duc et al. \cite{duc2021self} introduced an efficient self-knowledge distillation approach where the model learns from embedded feature representations of two different views of the same data, demonstrating feasibility for edge device deployment. Thus, Chi et al. \cite{chi2023prefallkd} proposed PreFallKD, specifically targeting pre-impact fall detection in resource-constrained environments.
Despite these advancements, most existing approaches rely on conventional knowledge distillation, focusing primarily on transferring class probabilities or feature-level information \cite{li2023object}. While these methods reduce model complexity and improve efficiency, they often fail to capture the higher-order relationships and spatial-temporal dependencies crucial for analyzing skeletal data in fall detection. This limitation highlights the potential of relational knowledge distillation (RKD) \cite{park2019relational}, which preserves structural and relational information in skeletal data, enabling more accurate fall detection by capturing intricate joint interactions, particularly during impact events.

\section{The Proposed Approach}\label{sec:proposed_approach}

We introduce our proposed DistillH-Mamba architecture for impact detection in fall events. This section first defines the specific problem addressed, followed by our methodology that combines hypergraph modeling with state-space models through relational knowledge distillation.

\begin{figure*}[ht]
\includegraphics[width=1\textwidth]{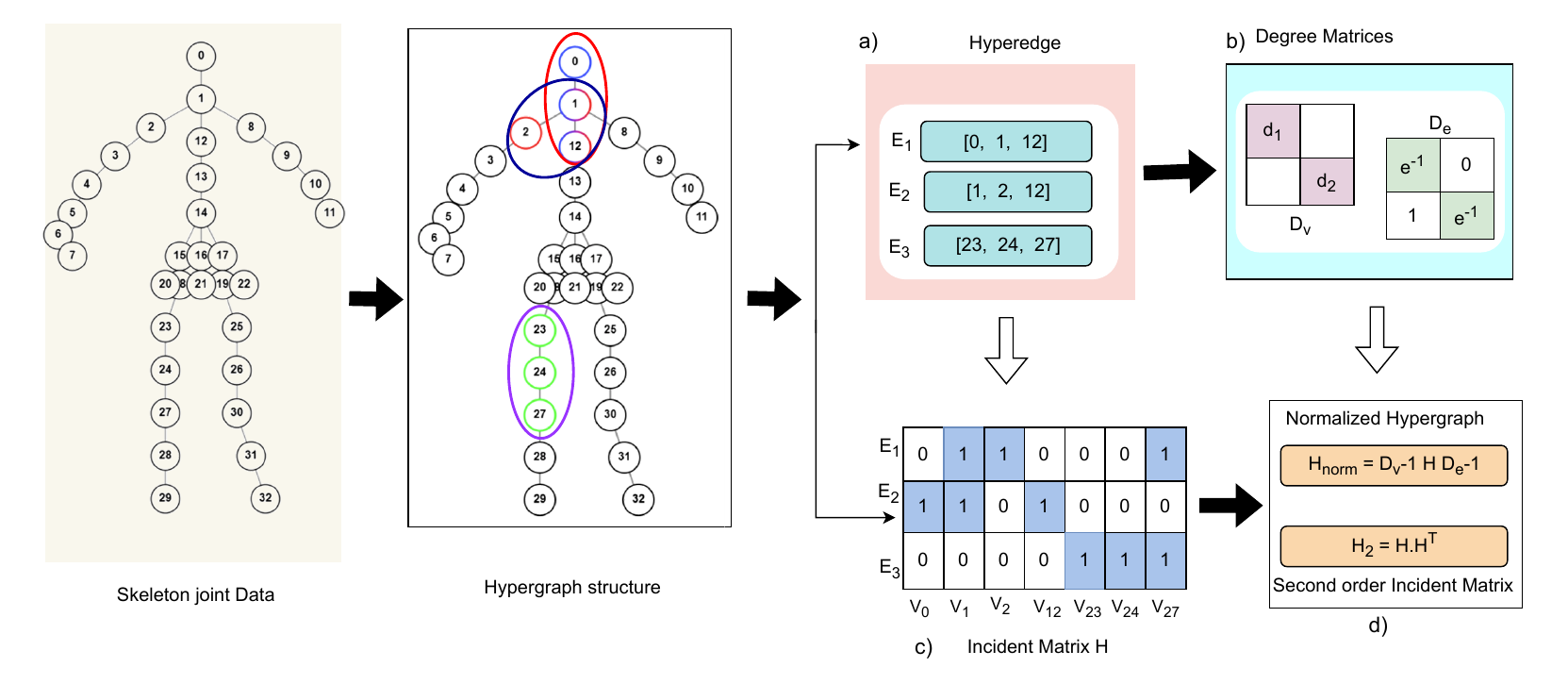}
\caption{Dual-Representation Hypergraph Construction Process for Skeletal Joint Data. 
a): Hyperedges (E) based on anatomical relationships, representing the set of vertices (V) and their connections. 
b): Degree matrices $\mathbf{D}_v$ (vertex degree) and $\mathbf{D}_e$ (hyperedge degree), capturing node and hyperedge connectivity. 
c): Incident matrix $\mathbf{H}$, where blue cells (value 1) indicate a vertex is in a hyperedge, and white cells (value 0) indicate no connection. 
d): Normalized hypergraph Laplacian $\mathbf{H}_{norm} = \mathbf{D}_v^{-1} \mathbf{H} \mathbf{D}_e^{-1}$ and second-order incident matrix $\mathbf{H}_2 = \mathbf{H}\mathbf{H}^T$.}
\label{hypergraph_construction}
\end{figure*}

\subsection{Problem Statement}

Fall detection systems aim to identify whether a fall event has occurred, effective medical intervention requires precise detection of the impact moment the critical point where an individual makes contact with the ground, potentially causing injury. Accurately identifying this moment can significantly reduce response time and improve healthcare outcomes for elderly individuals.
Given a skeletal motion sequence $S = \{s_1, s_2, ..., s_n\}$ where each $s_i$ represents the 3D joint positions at time $i$, we define the impact detection problem as identifying the specific frame $s_k$ where impact occurs. For each frame at time step $t$ (where $t \in \{1,2,...,n\}$), the skeleton is represented by a set of $J$ joints in 3D space. We denote the position of the $i$-th joint (where $i \in \{1,2,...,J\}$) at time $t$ as $j_i^t \in \mathbb{R}^3$, which is a 3D vector containing the $(x,y,z)$ coordinates of that joint. Therefore, a complete skeleton frame at time $t$ can be represented as $s_t = \{j_1^t, j_2^t, ..., j_J^t\}$.
This formulation differs significantly from traditional fall detection, which typically provides a binary classification for the entire sequence. Our approach addresses the challenge of impact detection within confirmed fall events, focusing on identifying the precise moment when the body hits the ground characterized by distinctive patterns in skeletal configuration and motion dynamics. By accurately pinpointing this critical moment, our system enables more timely intervention and better assessment of fall severity.

\subsection{Hypergraph Modeling for Skeletal Data}
% \color{black}  % or \normalcolor

Traditional graph-based approaches for representing human skeletons model joints as nodes with edges connecting physically adjacent joints. While this captures basic structural information, it fails to model the complex higher-order relationships between multiple joints during falls, particularly at impact moments. To address this limitation, we propose a dual-representation hypergraph approach specifically designed to capture complex joint dynamics during impact.

\subsubsection{First-order Hypergraph Construction}

A hypergraph $\mathcal{G} = (\mathcal{V}, \mathcal{E})$ consists of vertices $\mathcal{V} = \{v_0, v_1, ..., v_{J-1}\}$ representing $J$ skeleton joints, and hyperedges $\mathcal{E} = \{e_0, e_1, ..., e_{E-1}\}$ where each $e_j \subseteq \mathcal{V}$ connects two or more vertices. This enables modeling relationships between joint groups exhibiting coordinated movement during impacts.
As shown in Figure \ref{hypergraph_construction}, we transform skeletal data into a hypergraph by defining hyperedges based on anatomical structure and functional relationships. We construct an incidence matrix $\mathbf{H} \in \mathbb{R}^{J \times E}$ where $H_{i,j}=1$ if vertex $v_i$ belongs to hyperedge $e_j$, and 0 otherwise. Figure \ref{hypergraph_construction}.(c) illustrates this process, showing example hyperedges like $E_1 = \{0, 1, 12\}$, $E_2 = \{1, 2, 12\}$, and $E_3 = \{23, 24, 27\}$ from Figure \ref{hypergraph_construction}.(a) and their corresponding representation in the incident matrix $\mathbf{H}$.
For computational efficiency, we normalize this matrix using degree matrices, as
shown in Figure \ref{hypergraph_construction}.(b). Let $\mathbf{D}_v \in \mathbb{R}^{n \times n}$ be a matrix where $D_v(i,i) = \sum_{j=0}^{m-1} H_{i,j}$ represents hyperedges containing vertex $i$, and $\mathbf{D}_e \in \mathbb{R}^{m \times m}$ be a matrix where $D_e(j,j) = \sum_{i=0}^{n-1} H_{i,j}$ represents vertices in hyperedge $j$. The normalized hypergraph Laplacian is computed as $\mathbf{H}_{norm} = \mathbf{D}_{v}^{-1} \mathbf{H} \mathbf{D}_{e}^{-1}$ as shown in Figure \ref{hypergraph_construction}.(d).

\subsubsection{Second-order Hypergraph Representation}

While the first-order hypergraph captures direct relationships between joints within defined hyperedges, we extend our approach by computing a second-order incident matrix $\mathbf{H}_2 = \mathbf{H}\mathbf{H}^T$, as shown in Figure \ref{hypergraph_construction}.(d). This creates a vertex-vertex incident matrix where each entry $(i,j)$ indicates the number of hyperedges that joints $i$ and $j$ share. 
This second-order representation captures higher-level joint associations that might not be directly connected in the original hypergraph but exhibit functional relationships through shared hyperedges. For instance, joints appearing together in multiple hyperedges such as joints 1 and 12 appearing in both $E_1$ and $E_2$ in Figure \ref{hypergraph_construction}.(a) will have a stronger connection in $\mathbf{H}_2$, indicating their functional correlation during impact motion.
This representation is particularly important for modeling impact events, as ground contact creates complex force transmission patterns through the body that affect joints far beyond the initial contact point. By processing these complementary representations in parallel pathways, our model simultaneously captures direct joint relationships (first-order) and complex joint interactions (second-order), providing a comprehensive understanding of the distinctive coordination patterns that occur at the precise moment of ground contact.

\begin{figure}[hhttp]
\centering
\includegraphics[width=9.2cm]{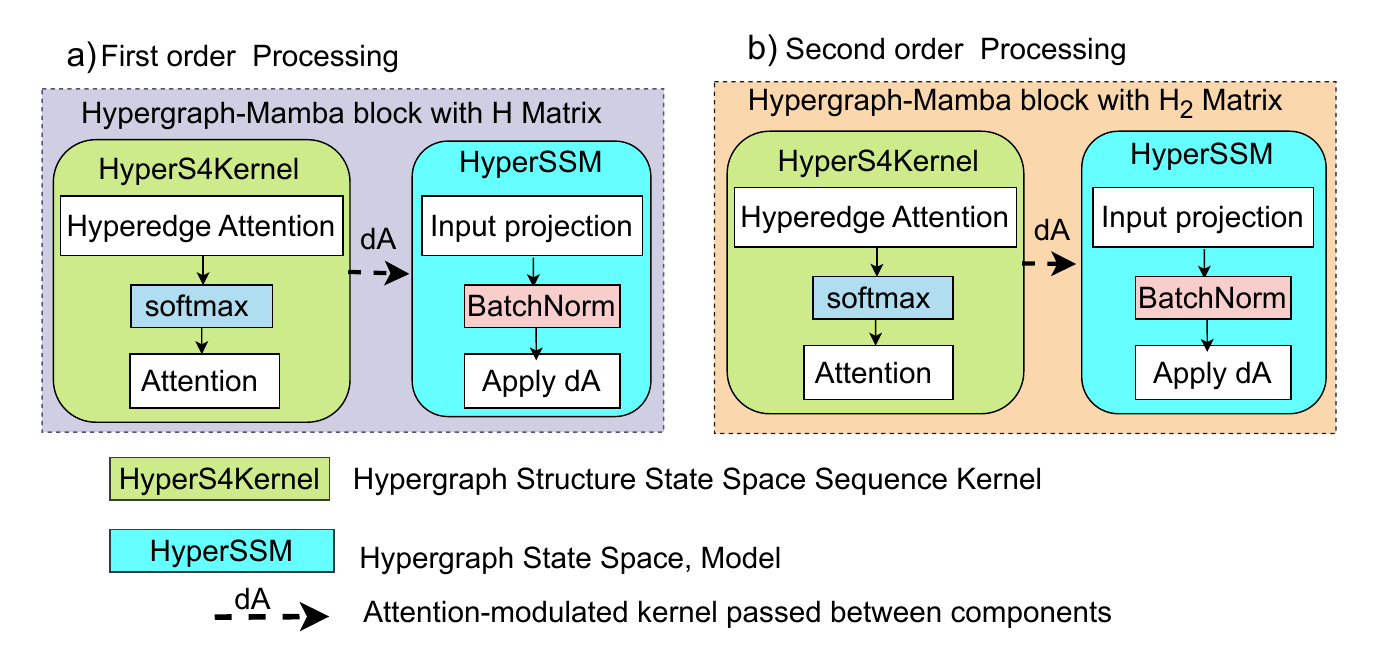}
\caption{Hypergraph-Mamba Block: a) First-order processing pathway using hypergraph incidence matrix $\mathbf{H}$, b) Second-order processing pathway using derived matrix $\mathbf{H}_2$}
\label{hypermamba}
\end{figure}

\subsection{Hypergraph-Mamba: Integrating Hypergraphs with State-Space Models}

To address the impact detection problem formulated previously, we propose the Hypergraph-Mamba block. As shown in Figure \ref{hypermamba}, this block integrates hypergraph representation with state-space models to precisely identify impact within fall events, capturing both the spatial relationships between joints and their temporal dynamics during the critical moment when the body makes contact with the ground.

\begin{figure*}[ht]
\centering
\includegraphics[width=1\textwidth]{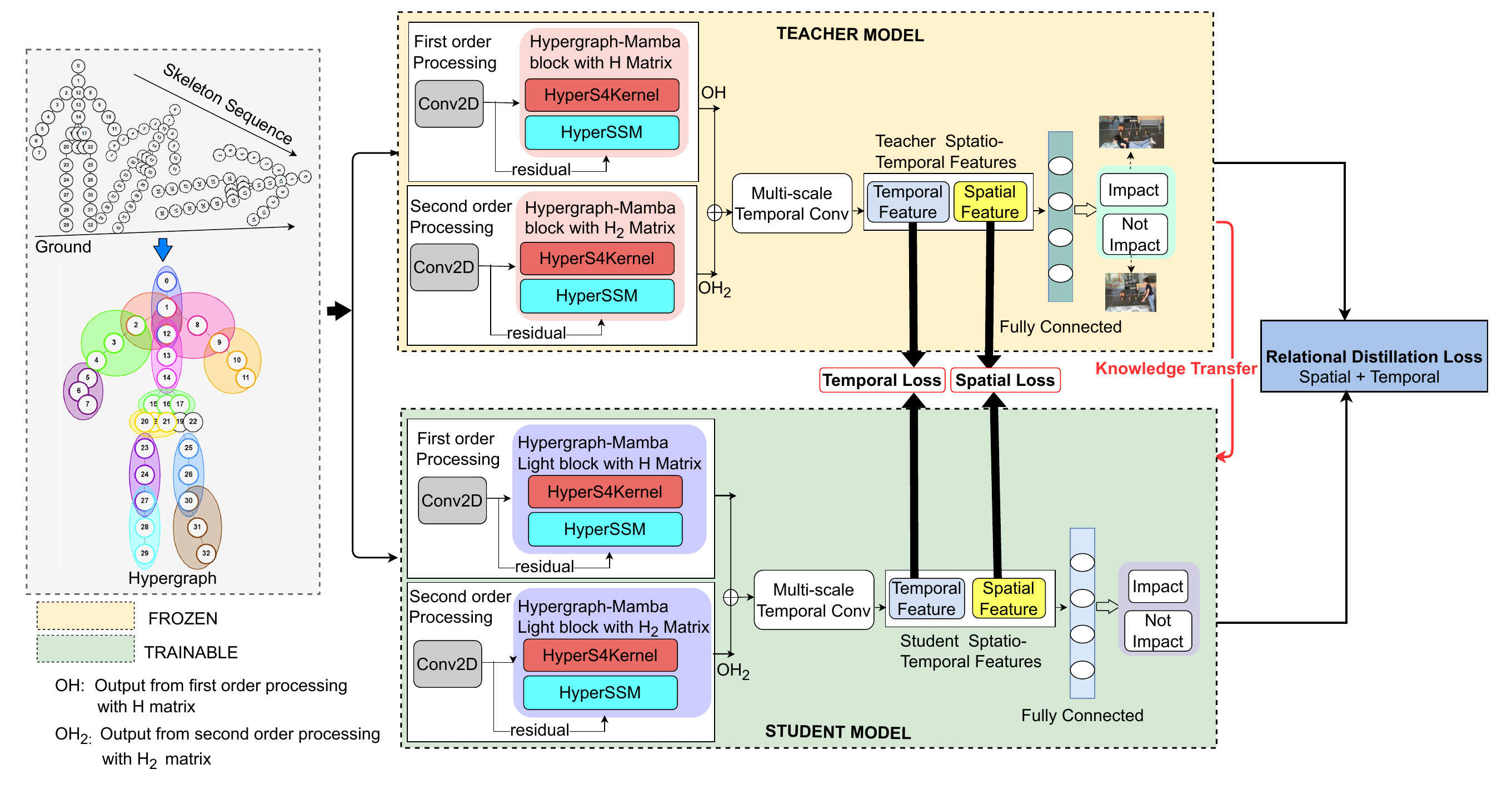}
\caption{Flowchart of the proposed DistillH-Mamba Architecture}
\label{pipeline}
\end{figure*}

\subsubsection{State-Space Models and Mamba Architecture}

State-space models (SSMs) provide an efficient alternative with linear-time complexity, particularly well-suited for modeling the distinctive temporal patterns of impact events while maintaining computational efficiency. The SSM is described by the following differential equations:

\begin{equation}
\frac{dx(t)}{dt} = \mathbf{A}x(t) + \mathbf{B}u(t) \quad y(t) = \mathbf{C}x(t) + \mathbf{D}u(t)
\end{equation}

where $x(t) \in \mathbb{R}^N$ is the hidden state at time $t$, $u(t) \in \mathbb{R}^M$ is the input at time $t$, $y(t) \in \mathbb{R}^M$ is the output at time $t$, $\mathbf{A} \in \mathbb{R}^{N \times N}$ is the state transition matrix, $\mathbf{B} \in \mathbb{R}^{N \times M}$ is the input projection matrix, $\mathbf{C} \in \mathbb{R}^{M \times N}$ is the output projection matrix, and $\mathbf{D} \in \mathbb{R}^{M \times M}$ is the feedforward matrix.

For implementation, the continuous system is discretized using a step size $\Delta$:

\begin{equation}
x_k = \bar{\mathbf{A}} x_{k-1} + \bar{\mathbf{B}} u_k \quad y_k = \mathbf{C}x_k + \mathbf{D}u_k
\end{equation}

where $\bar{\mathbf{A}} = e^{\Delta\mathbf{A}} \in \mathbb{R}^{N \times N}$ is the discretized state transition matrix, $\bar{\mathbf{B}} = (\Delta\mathbf{A})^{-1}(e^{\Delta\mathbf{A}} - \mathbf{I})(\Delta\mathbf{B}) \in \mathbb{R}^{N \times M}$ is the discretized input projection matrix, and $\mathbf{I} \in \mathbb{R}^{N \times N}$ is the identity matrix. State-space models (SSMs) provide an efficient alternative with linear-time complexity $O(l)$, particularly well-suited for modeling the distinctive temporal patterns of impact events while maintaining computational efficiency compared to the quadratic $O(l^2)$ complexity of traditional RNN and transformer-based approaches used in skeleton-based fall detection.

Building on these foundations, the Mamba architecture extends standard SSMs by introducing a selective mechanism that makes state-space parameters input-dependent:

\begin{equation}
\mathbf{A}_k = A(u_k) \quad \mathbf{B}_k = B(u_k)
\end{equation}

where $A$ and $B$ are learned functions that map inputs to parameters, dynamically adjusting how information flows through the network based on input features.

\subsubsection{Proposed Hypergraph-Mamba Block}

The proposed Hypergraph-Mamba block addresses three key challenges in impact detection that existing methods often struggle to handle effectively. First, impact events involve complex coordination across multiple joints, where forces propagate through interconnected anatomical structures. This necessitates modeling higher-order relationships, beyond simple pairwise joint connections. Second, accurate impact detection depends on high temporal sensitivity to capture sudden changes in motion that happen very quickly, while still keeping track of the entire fall sequence. Third, real-time deployment imposes strict computational constraints, requiring models to be both efficient and accurate. Our architecture is driven by the complementary strengths of hypergraphs and state-space models. Hypergraphs are well suited for representing simultaneous multi-joint interactions during an impact, while state-space models particularly the Mamba architecture are effective at capturing rapid temporal dynamics with linear computational complexity, which is essential for real-time processing. As illustrated in Figure~\ref{hypermamba}, the Hypergraph-Mamba block operates through two parallel pathways: a first-order processing pathway using the original hypergraph incidence matrix $\mathbf{H}$, and a second-order pathway using the derived matrix $\mathbf{H}_2$. Each pathway includes two core components HyperS4Kernel and HyperSSM which interact via attention-modulated kernel passing. This dual-pathway design enables the model to capture both explicit anatomical structures and emergent coordination patterns that are critical for reliable impact detection.

\paragraph{HyperS4Kernel: Hypergraph Structure State Space Sequence Kernel}

The HyperS4Kernel serves as the spatial relationship encoder, transforming hypergraph structure into attention weights that guide temporal processing. It receives skeletal joint features and computes hypergraph-aware attention that respects anatomical relationships, producing an attention-modulated kernel that encodes which joint relationships are most relevant for the current temporal context. This design ensures that temporal processing focuses on biomechanically meaningful joint coordination patterns rather than arbitrary feature combinations.

Given an input feature sequence $\mathbf{X} \in \mathbb{R}^{T \times J \times C}$, where $T$ is the sequence length, $J$ is the number of skeletal joints, and $C$ is the feature dimension, the HyperS4Kernel computes state transitions that respect the underlying joint relationships through a hypergraph attention mechanism:

\begin{equation}
\boldsymbol{\alpha} = \text{softmax}(\mathbf{W}_{\text{attn}})
\end{equation}

where $\mathbf{W}_{\text{attn}} \in \mathbb{R}^{J \times E}$ is a learnable parameter matrix representing attention weights, $\boldsymbol{\alpha} \in \mathbb{R}^{J \times E}$ is the attention weight matrix, $J$ is the number of skeletal joints, and $E$ is the number of hyperedges.

These attention scores are modulated by the respective incidence matrix:
\begin{equation}
\mathbf{A}_{\text{hyper}} = \boldsymbol{\alpha} \odot \mathbf{H}
\end{equation}

where $\mathbf{H} \in \mathbb{R}^{J \times E}$ is the hypergraph incidence matrix, $\mathbf{A}_{\text{hyper}} \in \mathbb{R}^{J \times E}$ is the hypergraph attention matrix, and $\odot$ represents element-wise multiplication.
The hypergraph attention is aggregated along the hyperedge dimension:
\begin{equation}
\mathbf{A}_{\text{node}} = \sum_{j=0}^{E-1} \mathbf{A}_{\text{hyper},j}
\end{equation}

where $\mathbf{A}_{\text{node}} \in \mathbb{R}^{J}$ is the aggregated node attention vector. The final attention-modulated kernel is computed as:
\begin{equation}
\mathbf{dA} = \text{softmax}(\mathbf{K}) \odot \mathbf{A}_{\text{node}}
\end{equation}

where $\mathbf{K} \in \mathbb{R}^{J \times P \times P}$ is a learnable kernel matrix and $\mathbf{dA} \in \mathbb{R}^{J \times P \times P}$ is the final attention-modulated kernel.

\paragraph{HyperSSM: Hypergraph State Space Model}

The HyperSSM component functions as the temporal dynamics processor, using the attention-modulated kernel from HyperS4Kernel to guide selective state updates. The selective gating mechanism enables the model to maintain stable representations during normal motion while rapidly adapting to sudden changes characteristic of impact events. This component interaction creates a feedback loop where spatial joint relationships inform temporal processing, while temporal context influences which spatial relationships receive attention. The HyperSSM component receives the attention-modulated kernel $\mathbf{dA} \in \mathbb{R}^{J \times P \times P}$ from the HyperS4Kernel along with the input feature sequence $\mathbf{X} \in \mathbb{R}^{T \times J \times C}$. It processes features through hypergraph-aware state transitions:
\begin{equation}
\mathbf{h}_t = \phi(\mathbf{dA} \cdot \mathbf{h}_{t-1} + \mathbf{W}_{\text{in}}\mathbf{x}_t)
\end{equation}

where $\mathbf{h}_t \in \mathbb{R}^{J \times P}$ is the hidden state at time $t$, $\mathbf{x}_t \in \mathbb{R}^{J \times c_{\text{in}}}$ is the input feature, and $\mathbf{W}_{\text{in}} \in \mathbb{R}^{P \times c_{\text{in}}}$ is an input projection matrix.

The model implements a selective gating mechanism that effectively detects sudden changes in joint dynamics:
\begin{equation}
\mathbf{g}_t = \sigma(\mathbf{W}_g \cdot \mathbf{x}_t + \mathbf{b}_g)
\end{equation}

\begin{equation}
\tilde{\mathbf{h}}_t = \mathbf{g}_t \odot \mathbf{h}_t + (1 - \mathbf{g}_t) \odot \mathbf{h}_{t-1}
\end{equation}

where $\mathbf{g}_t \in \mathbb{R}^{J \times P}$ is a gating function controlling information flow.

The output features are then computed:
\begin{equation}
\mathbf{y}_t = \mathbf{W}_{\text{out}}\tilde{\mathbf{h}}_t
\end{equation}

where $\mathbf{W}_{\text{out}} \in \mathbb{R}^{c_{\text{out}} \times P}$ is an output projection matrix.
By incorporating hypergraph-aware attention, HyperSSM selectively focuses on joint relationship patterns that exhibit characteristic changes during ground contact, enabling rapid detection of impact onset while maintaining context about the preceding fall sequence necessary for accurate classification.
The overall output of the Hypergraph-Mamba processing can be expressed as:
\begin{equation}
\mathbf{Z}^t = \text{Reshape}(\mathbf{Y}^t) \mathbf{W}_o \in \mathbb{R}^{J \times G}
\end{equation}

where $\mathbf{Y}^t$ represents the spatio-temporal features at time step $t$, and $\mathbf{Z}^t$ is the final impact-sensitive joint representation. This integration of dual-representation hypergraphs with selective state-space modeling effectively detects the abrupt changes in joint dynamics that occur at the moment of ground impact during falls.

\begin{table*}[http]
\centering
% \caption{Performance comparison of multi-scale temporal convolution architectures for impact detection.}
\caption{{Performance comparison of multi-scale temporal convolution architectures for impact detection}}
\label{tab:multiscale_validation}
\begin{tabular}{lcccccc}
\hline
\textbf{Architecture} & \textbf{Processing Strategy} & \textbf{Accuracy (\%)} & \textbf{Precision (\%)} & \textbf{Specificity (\%)} & \textbf{F1-Score (\%)} & \textbf{Inference (ms)} \\
\hline
Parallel (\textbf{Ours}) & Independent multi-scale & \textbf{97.38} & \textbf{96.38} & \textbf{99.73} & \textbf{97.51} & \textbf{24.8} \\
Single-Scale & Single $k=15$ kernel & 90.12 & 92.31 & 91.33 & 90.67 & 28.5 \\
Serial & Sequential $k=9 \rightarrow k=15 \rightarrow k=20$ & 90.00 & 93.98 & 93.50 & 90.36 & 47.8 \\
\hline
\end{tabular}
\end{table*}
\subsection{Multi-scale Temporal Convolutions for Impact Characterization}
\begin{figure}[t]
\centering
\includegraphics[width=9cm]{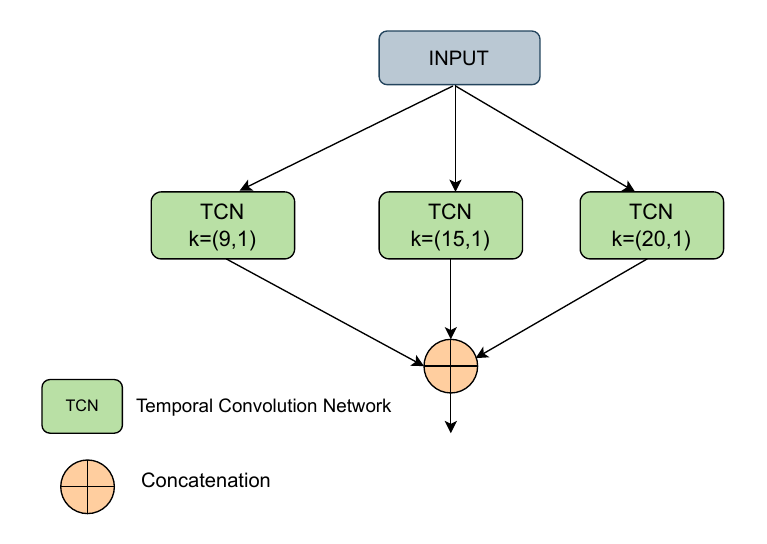}
\caption{Multi-scale Temporal Convolution Network with Parallel Processing. Independent temporal convolutions with different kernel sizes process the same input simultaneously to capture multi-scale temporal patterns for impact detection.}
\label{tcn}
\end{figure}

Impact events during falls exhibit characteristic temporal patterns at different time scales, from rapid ground contact to force transmission through the skeletal structure. To capture these multi-scale patterns, we employ a parallel temporal convolution architecture as illustrated in Figure~\ref{tcn}. 
Our design processes the combined hypergraph features through three independent temporal convolution branches operating simultaneously on the same input, with each branch employing 2D convolutions with different kernel sizes along the temporal dimension:
% {Impact events during falls exhibit characteristic temporal patterns at different time scales. To capture these multi-scale patterns, we employ a series of temporal convolutions with increasing receptive fields as shown in Figure \ref{tcn}. Each TCN branch in our multi-scale architecture is implemented using 2D convolutions. Following the parallel Hypergraph-Mamba blocks, we apply three Conv2D operations with progressively larger kernel sizes along the temporal dimension:}
\begin{equation}
\mathbf{Z}_k = \sigma(\text{Conv2D}(\mathbf{F}, (k, 1), \text{padding='same'})) 
\end{equation}

where $k \in \{9, 15, 20\}$ represents the temporal kernel size, $\mathbf{F}$ denotes the input feature map from the hypergraph processing, and $\sigma$ is the ReLU activation function. The kernel shape $(k, 1)$ ensures temporal processing while preserving the joint spatial dimension.
Each temporal scale captures specific impact characteristics: the fine-scale kernel ($k=9$) detects  patterns at the precise moment of ground contact, the medium-scale kernel ($k=15$) identifies transitional movement patterns and force propagation during impact, and the coarse-scale kernel ($k=20$) provides broader temporal context around the impact event.
% {where $k \in \{9, 15, 20\}$ represents the temporal kernel size, $\mathbf{F}$ is the input feature map, and $\sigma$ is a ReLU activation function. The kernel shape $(k, 1)$ enables the convolutions to operate along the temporal dimension while preserving the joint dimension.
% The smaller kernel ($k=9$) captures rapid pattern at the precise moment of impact. The medium kernel ($k=15$) identifies transitional movements immediately following ground contact. The larger kernel ($k=20$) extracts broader context of posture changes both before and after impact.
% These multi-scale features are then concatenated to form a comprehensive impact representation.
% } 
The outputs from all three scales are concatenated along the channel dimension, creating an enriched feature representation that preserves the temporal specialization of each scale while enabling the model to learn optimal combinations of multi-scale patterns. This parallel architecture fundamentally differs from sequential approaches that process temporal scales consecutively, potentially creating information bottlenecks where early-scale features are progressively transformed and lost at each processing stage. This information preservation is critical, as the precise identification of ground contact requires simultaneous analysis of rapid changes ($k=9$), force propagation through the kinematic chain ($k=15$), and the broader contextual dynamics of the fall sequence ($k=20$). The independent parallel processing ensures that each temporal scale preserves its characteristic patterns without interference from other scales, while the concatenation operation maintains all temporal information in a unified representation that captures the complete multi-scale dynamics of impact events

\subsubsection{Architectural Validation}

To validate the effectiveness of our parallel multi-scale design against alternative temporal processing strategies, we conducted comprehensive experiments comparing three distinct architectural approaches, as presented in Table~\ref{tab:multiscale_validation}. These comparisons are essential to demonstrate that our parallel approach provides superior information preservation and processing efficiency compared to conventional single-scale or sequential multi-scale methods.
Our parallel architecture achieves superior performance with 97.38\% accuracy, representing significant improvements of 7.26\% over single-scale and 7.38\% over serial alternatives. The exceptional 99.73\% specificity is particularly crucial for emergency response systems, as it minimizes false alarms that could lead to caregiver alert fatigue and reduced system trust. Additionally, the 24.8ms inference time enables real-time processing capabilities essential for time-critical healthcare monitoring applications.
The performance gains demonstrate that preserving all temporal scales simultaneously provides substantially richer feature representations compared to approaches that compress temporal information into a single scale or sequentially process multi-scale features with potential information loss. The parallel approach maintains distinct temporal features for each scale that collectively encode the complete temporal signature of impact events, whereas the serial approach progressively transforms and potentially degrades early-scale information through successive convolution operations. These empirical results validate our parallel multi-scale approach as optimal for precise impact detection.

\subsection{DistillH-Mamba Architecture for Fast Impact Detection}

While our Hypergraph-Mamba model effectively captures the complex dynamics of fall impacts, its computational complexity and memory requirements may limit deployment on resource-constrained devices. To address this challenge, we employ Relational Knowledge Distillation (RKD) to create a lightweight architecture that preserves essential spatial-temporal relationships while significantly reducing computational demands. The overall architecture of our proposed DistillH-Mamba framework is illustrated in Figure \ref{pipeline}.

\begin{table}[http]
\centering

\caption{Parameters of Teacher and Student DistillH-Mamba Models}
\label{tab1}
\begin{tabular}{lcc}
\toprule
\textbf{Component} & \textbf{Teacher} & \textbf{Student} \\
\midrule
\multicolumn{3}{l}{\textit{Initial Feature Extraction}} \\
\midrule
Initial Conv Filters & 64 & 48 \\
X-branch Conv Filters & 26 & 26 \\
Y-branch Conv Filters & 33 & 33 \\
\midrule
\multicolumn{3}{l}{\textit{Hypergraph-Mamba Blocks}} \\
\midrule
First-order Hypergraph-Mamba d\_state & 128 & 64 \\
Second-order Hypergraph-Mamba d\_state & 64 & 32 \\
Dropout Rate & 0.25 & 0.25 \\
\midrule
\multicolumn{3}{l}{\textit{Temporal Processing}} \\
\midrule
Conv\_z1 Filters & 16 & 12 \\
Conv\_z2 Filters & 16 & 12 \\
Conv\_z3 Filters & 16 & 12 \\
Temporal Kernel Sizes & 9, 15, 20 & 9, 15, 20 \\
\midrule
\end{tabular}
\end{table}

\subsubsection{Teacher-Student Distillation Framework}

Our RKD Framework consists of a teacher-student model where the teacher is the full Hypergraph-Mamba model that captures complex higher-order relationships between joints and temporal dynamics but is computationally intensive. The student model employs a simplified architecture with reduced dimensionality and optimized components as described in Table \ref{tab1}.
Despite these reductions in model capacity, the student model maintains the dual-pathway hypergraph structure, preserving the ability to model both direct joint relationships through the first-order incidence matrix $\mathbf{H}$ and complex interactions through the second-order representation $\mathbf{H}_2$.
Algorithm \ref{alg:distillh_mamba} presents the complete training procedure for our DistillH-Mamba framework, encompassing both teacher model training and the relational knowledge distillation process for the student model.

\begin{algorithm}[h]
\caption{DistillH-Mamba Training Process}
\label{alg:distillh_mamba}
\begin{algorithmic}[1]
\Require {Skeletal sequences $S$, ground truth labels $Y$, hyperparameters $\alpha$, $\lambda_{spatial}$, $\lambda_{temporal}$}
\Ensure {Trained student model $\theta_S$}
 
\State {\textbf{Phase 1: Teacher Training}}
\State {Initialize teacher model $\theta_T$ with full capacity (Table \ref{tab1})}
\State {Construct dual hypergraphs $\mathbf{H}$ and $\mathbf{H}_2$}
\For{{epoch in training\_epochs}}
    \State {Extract spatio-temporal features using Hypergraph-Mamba blocks}
    \State {Apply multi-scale temporal convolutions (Eq. 13)}
    \State {Compute classification loss $\mathcal{L}_{BCE}$}
    \State {Update $\theta_T$ using gradient descent}
\EndFor
 
\State {\textbf{Phase 2: Student Training with RKD}}
\State {Initialize student model $\theta_S$ with reduced capacity (Table \ref{tab1})}
\For{{epoch in distillation\_epochs}}
    \State 
    {Forward pass through both teacher and student models}
    \State 
    {Compute spatial loss: $\mathcal{L}_{spatial} = \|\mathbf{F}^T - \mathbf{F}^S\|_F$ (Eq. 14)}
    \State 
    {Compute temporal loss: $\mathcal{L}_{temporal} = \|\Delta\mathbf{F}^T - \Delta\mathbf{F}^S\|_F$ (Eq. 15)}
    \State {Compute total loss: $\mathcal{L}_{total} = (1-\alpha)\mathcal{L}_{BCE} + \alpha(\lambda_{spatial}\mathcal{L}_{spatial} + \lambda_{temporal}\mathcal{L}_{temporal})$ (Eq. 16)}
    \State {Update $\theta_S$ using gradient descent}
\EndFor
\State \Return {$\theta_S$}
\end{algorithmic}
\end{algorithm}

\subsubsection{Relational Knowledge Distillation}

While conventional knowledge distillation approaches primarily transfer class probabilities or feature representations, they often fail to preserve structured relationships crucial for skeletal data analysis \cite{moslemi2024survey}. For impact detection, spatial-temporal relationships between joints are fundamental determinants of classification accuracy. Therefore, we employ relational knowledge distillation (RKD) to explicitly preserve these structural relationships during model compression, maintaining the model's capacity to identify distinctive joint configurations that characterize impact moments.
We define two key relational distillation objectives:
Firstly, we preserve spatial relationships between joints using a hypergraph relation consistency loss:
\begin{equation}
\mathcal{L}_{\text{spatial}} = \| \mathbf{F}^T_{\text{hgcn-x}} - \mathbf{F}^S_{\text{hgcn-x}} \|_F + \| \mathbf{F}^T_{\text{hgcn-y}} - \mathbf{F}^S_{\text{hgcn-y}} \|_F
\end{equation}

where $\mathbf{F}^T$ and $\mathbf{F}^S$ are the hypergraph features from the teacher and student models respectively, with $\mathbf{F}_{\text{hgcn-x}}$ representing features from the first-order hypergraph pathway (processing the original incidence matrix $\mathbf{H}$) and $\mathbf{F}_{\text{hgcn-y}}$ representing features from the second-order hypergraph pathway (processing the derived incident matrix $\mathbf{H}_2 = \mathbf{H}\mathbf{H}^T$). The Frobenius norm ($\|\cdot\|_F$) effectively captures the overall structural similarity between feature representations.
Secondly, we maintain temporal dynamics through a temporal relation distillation loss:
\begin{equation}
\mathcal{L}_{\text{temporal}} = \| \Delta \mathbf{F}^T_{\text{hgcn-x}} - \Delta \mathbf{F}^S_{\text{hgcn-x}} \|_F + \| \Delta \mathbf{F}^T_{\text{hgcn-y}} - \Delta \mathbf{F}^S_{\text{hgcn-y}} \|_F
\end{equation}

where $\Delta \mathbf{F} = \mathbf{F}_{t} - \mathbf{F}_{t-1}$ represents the temporal difference between consecutive frames. This temporal loss is crucial for detecting the sudden changes that characterize impact moments, ensuring the student model preserves the teacher's sensitivity to rapid changes in joint dynamics during ground contact.
The overall distillation loss combines these relational losses with the standard binary cross-entropy classification loss:
\begin{equation}
\mathcal{L}_{\text{total}} = (1-\alpha)\mathcal{L}_{\text{BCE}} + \alpha(\lambda_{\text{spatial}} \mathcal{L}_{\text{spatial}} + \lambda_{\text{temporal}} \mathcal{L}_{\text{temporal}})
\end{equation}

where $\alpha$ controls the overall importance of distillation relative to direct supervision, while $\lambda_{\text{spatial}}$ and $\lambda_{\text{temporal}}$ balance the contribution of the spatial and temporal relational components.

\section{Experimentation and Results}

We evaluate our DistillH-Mamba architecture for impact fall detection, analyzing how each component such as hypergraph structure, Mamba architecture, and knowledge distillation contributes to overall performance. Our experiments assess both detection accuracy and computational efficiency, demonstrating the effectiveness of relational knowledge distillation in preserving critical capabilities while significantly reducing resource requirements.

\subsection{Data Description and Ethical Considerations}

Obtaining real-world data from the elderly is still challenging due to privacy issues. Thus, to evaluate our framework, we used two publicly available datasets: the 3D Skeletons UP-Fall Dataset \cite{koffi2025improved} and the UMAFall dataset \cite{casilari2017umafall}.

\textbf{3D Skeletons UP-Fall Dataset} is a dataset built upon the original UP-Fall dataset \cite{martinez2019up} by providing richer, more detailed 3D skeleton data specifically tailored for detecting impacts during fall events. This dataset includes data from five subjects performing five fall scenarios, along with activities of daily living (ADLs). These scenarios simulate real-world fall situations, ensuring diverse and realistic representations of fall dynamics. The 3D skeleton data was obtained using MediaPipe Pose Estimation, extracting 33 joint positions represented as 3D coordinates (x, y, z) from each frame. The dataset composition is detailed in Table~\ref{tab:upfall_activities}.

\begin{table}[htbp]
\centering
\small
\caption{Description of Activities in the 3D Skeletons UP-Fall Dataset}
\label{tab:upfall_activities}
\begin{tabular}{ccl}
\toprule
\textbf{Activity ID} & \textbf{Category} & \textbf{Description} \\
\midrule
1 & Fall & Forward fall \\
2 & Fall & Backward fall \\
3 & Fall & Lateral fall to the right \\
4 & Fall & Lateral fall to the left \\
5 & Fall & Fall from sitting \\
\midrule
6 & ADL & Walking \\
7 & ADL & Standing \\
8 & ADL & Sitting \\
9 & ADL & Picking up an object \\
10 & ADL & Jumping \\
\bottomrule
\end{tabular}
\end{table}

\textbf{UMAFall dataset} is a publicly multimodal dataset \cite{casilari2017umafall} acquired through systematic emulation of predefined Activities of Daily Life (ADLs) and falls. The dataset contains three distinct types of emulated falls on a mattress: lateral, frontal, and backward falls. These types of falls correspond to similar types of falls in the D Skeletons UP-
Fall Dataset, along with various activities of daily living. This similarity allows for direct comparison and evaluation of fall detection algorithms across different datasets, providing insights into the model's robustness and generalizability. The complete activity breakdown is presented in Table~\ref{tab:umafall_activities}.

\begin{table}[htbp]
\centering
\small
\caption{Activities Performed by Subjects in UMAFall Dataset}
\label{tab:umafall_activities}
\begin{tabular}{ccl}
\toprule
\textbf{Activity ID} & \textbf{Category} & \textbf{Description} \\
\midrule
1 & FALL & Falling forward \\
2 & FALL & Falling backward \\
3 & FALL & Falling lateral \\
\midrule
4 & ADL & Climbing stairs \\
5 & ADL & Raising the hands \\
6 & ADL & Walking \\
7 & ADL & Standing \\
8 & ADL & Sitting \\
9 & ADL & Jogging \\
10 & ADL & Body bending \\
11 & ADL & Lying down and getting up from bed \\
12 & ADL & Clapping hands \\
13 & ADL & Making a phone call \\
14 & ADL & Opening a door \\
15 & ADL & Jumping \\
16 & ADL & Laying down \\
\bottomrule
\end{tabular}
\end{table}

\textbf{Ethical Considerations:} Both public datasets were collected with informed consent from all participants by the original research teams, with appropriate ethical approvals from their respective institutions. All data were anonymized to protect participant privacy, and our use of these pre-existing datasets complied with ethical standards for secondary data analysis. Additionally, our self-created validation dataset was collected with informed consent from the participant, following appropriate ethical protocols for data collection and use.
\begin{table*}[t]
\centering
\caption{Ablation Study: Performance Comparison Across Model Variants using 3D Skeletons UP-Fall Dataset}
\label{ablation}
\begin{tabular}{lccccc}
\hline
\textbf{Algorithms} & \textbf{Accuracy} & \textbf{Precision} & \textbf{Recall} & \textbf{Specificity} & \textbf{Inference Time (s)} \\
\hline
GCN (Graph) & 92.00 & 91.43 & 95.59 & 84.38 & 2.364 × 10$^{-1}$ \\
HGCN (Hypergraph)  & 91.58 & 92.04 & 92.22 & 90.85 & 2.038 × 10$^{-1}$ \\
Hypergraph-Mamba (with only First order processing $\mathbf{H}$)  & 94.87 & 93.72 & 96.98 & 92.41 & 1.536 × 10$^{-1}$ \\
Hypergraph-Mamba (with only second order processing $\mathbf{H}_2$)  & 95.13 & 94.34 & 96.75 & 93.22 & 1.648 × 10$^{-1}$ \\
Hypergraph-Mamba (with both First and second order processing)  & 96.88 & 96.08 & 98.12 & 95.40 & 1.823 × 10$^{-1}$ \\
\textbf{Ours}(DistillH-Mamba-Student) & \textbf{97.38} & \textbf{96.38} & \textbf{98.84} & \textbf{99.73} & \textbf{0.478 × 10$^{-1}$} \\
\hline
\end{tabular}
\end{table*}

\subsection{Evaluation Metrics}

To comprehensively assess the performance of our impact detection system, we employ seven key metrics such as  accuracy, precision, specificity, F1-score, recall, and FLOPs, inference time. The $TP$ and $TN$  in Equations (17)–(21) are respectively the true positives and true negatives while $FP$ and $FN$ are the false positives and false negatives respectively. In the context of impact detection, $TP$ corresponds to correctly identified impact moments, $TN$ to correctly identified non-impact event, $FP$ to falsely detected impacts, and $FN$ to missed impact moments.

\begin{equation}
\text{accuracy} = \frac{\text{TP} + \text{TN}}{\text{TP} + \text{TN} + \text{FP} + \text{FN}} \label{eq:accuracy}
\end{equation}

Equation \ref{eq:accuracy} calculates the accuracy of the impact detection model in correctly identifying both impact and non-impact event within fall sequences. It assesses the model's overall correctness by considering the number of true positive (correctly identified impact moments) and true negative (correctly identified non-impact event) predictions relative to all instances, including false positives (non-impact event incorrectly classified as impacts) and false negatives (impact moments incorrectly classified as non-impacts).

\begin{equation}
\text{precision} = \frac{TP}{TP + FP} \label{eq:precision}
\end{equation}

Equation \ref{eq:precision} computes precision, this metric assessing the accuracy of positive predictions in distinguishing impact moments from non-impact event within fall sequences. It measures how well the model performs when predicting an impact moment (positive class).

\begin{equation}
\text{specificity} = \frac{TN}{TN + FP} \label{eq:specificity}
\end{equation}

Equation \ref{eq:specificity} calculates specificity, a crucial metric in impact detection within fall sequences. It measures the model's ability to correctly identify non-impact events by accurately recognizing true negatives among all actual non-impact instances.

\begin{equation}
\text{F1-score} = \frac{2 \cdot \text{precision} \cdot \text{sensitivity}}{\text{precision} + \text{sensitivity}} \label{eq:f1-score}
\end{equation}

Equation \ref{eq:f1-score} calculates the F1-score, a metric specifically designed for impact moment detection within fall sequences. It combines both precision and sensitivity to provide a balanced assessment of the model's performance in identifying impact moments accurately.

\begin{equation}
\text{recall} = \frac{TP}{TP + FN} \label{eq:recall}
\end{equation}

{Equation \ref{eq:recall} measures the model's ability to correctly identify positive class (impact moments) among all actual positive instances in fall sequences.}

\textbf{Inference Time:} Measures the computational efficiency of the model during impact detection deployment. This metric is crucial for real-time fall analysis systems where rapid impact identification is essential for timely response.

\textbf{FLOPs} (Floating Point Operations):  Quantifies the computational complexity of the impact detection model. This metric indicates the model's suitability for deployment on resource-constrained devices commonly used in fall monitoring systems.

\begin{table*}[t]
\centering
\caption{Ablation study on knowledge distillation effectiveness}
\label{tab:kd_ablation}
\begin{tabular}{lccccc}
\toprule
\textbf{Method} & \textbf{Accuracy (\%)} & \textbf{Precision (\%)} & \textbf{Recall (\%)} & \textbf{F1-Score (\%)} & \textbf{Specificity (\%)} \\
\midrule
Teacher (full capacity) & 96.88 & 96.08 & 98.12 & 96.40 & 92.14 \\
Student (no distillation) & 94.21 & 93.45 & 95.12 & 94.28 & 89.45 \\
Student + Standard KD & 95.67 & 94.82 & 96.34 & 95.57 & 91.23 \\
\textbf{Student + Relational KD (Ours)} & \textbf{97.38} & \textbf{96.38} & \textbf{98.84} & \textbf{97.51} & \textbf{99.73} \\
\midrule
\multicolumn{6}{l}{\textit{Improvements over baseline student model:}} \\
Standard KD improvement & +1.46 & +1.37 & +1.22 & +1.29 & +1.78 \\
Our RKD improvement & \textbf{+3.17} & \textbf{+2.93} & \textbf{+3.72} & \textbf{+3.23} & \textbf{+10.28} \\
\bottomrule
\end{tabular}
\end{table*}

\begin{table*}[htbp]
\centering
\caption{ Comparison with state-of-the-art skeleton-based methods on the UP-Fall dataset, highlighting impact detection capabilities within fall events.}
\begin{tabular}{lccccc}
\hline
\textbf{Method} & \textbf{Type} & \textbf{Fall} & \textbf{Impact} & \textbf{Accuracy (\%)} & \textbf{F1-score (\%)} \\
\hline
LSTM\cite{taufeeque2021multi} & Skeleton & \textcolor{green}{\checkmark} & \text{\textcolor{red}{\sffamily X}} & -- & 92.5 \\
CNNs\cite{galvao2021multimodal}& Skeleton & \textcolor{green}{\checkmark} & \text{\textcolor{red}{\sffamily X}} & \textbf{98.62} & 93.00 \\
SDFA\cite{zahan2022sdfa} & Skeleton & \textcolor{green}{\checkmark} & \text{\textcolor{red}{\sffamily X}} & 88.71 & - \\
ST-GCN\cite{yan2023skeleton} & Skeleton & \textcolor{green}{\checkmark} & \text{\textcolor{red}{\sffamily X}} & 97.27 & 83.57 \\
BERT\cite{ramirez2023bert} & Skeleton & \textcolor{green}{\checkmark} & \text{\textcolor{red}{\sffamily X}} & 81.14 & 80.95 \\
\hline
\textbf{(Ours)} DistillH-Mamba-Student  & \textbf{Skeleton} & \textbf{\textcolor{green}{\checkmark}} & \textbf{\textcolor{green}{\checkmark}} & 97.38 & \textbf{97.51} \\
\hline
\end{tabular}
\label{sota}
\end{table*}

\begin{table*}[htbp]
\centering
\caption{Comparison with recent state-of-the-art fall detection methods across different datasets.}
\begin{tabular}{lcccc}
\hline
\textbf{Method} & \textbf{Fall} & \textbf{Impact} & \textbf{Accuracy (\%)} & \textbf{F1-score (\%)} \\
\hline
Transformer-based fall detection \cite{nunez2024transformer} & \textcolor{green}{\checkmark} & \text{\textcolor{red}{\sffamily X}} & 96.67 & 82.24 \\
MSTSK-GFFk\cite{amsaprabhaa2023multimodal} & \textcolor{green}{\checkmark} & \text{\textcolor{red}{\sffamily X}} & 96.53 & 96.95 \\
FL-FD\cite{qi2023fl} & \textcolor{green}{\checkmark} & \text{\textcolor{red}{\sffamily X}} & \textbf{99.92} & -- \\
CNN+LSTM\cite{inturi2023novel} & \textcolor{green}{\checkmark} & \text{\textcolor{red}{\sffamily X}} & 98.59& 92.47 \\
Fall-Mamba\cite{zhang2025fall} & \textcolor{green}{\checkmark} & \text{\textcolor{red}{\sffamily X}} & 99.63 & - \\
\hline
\textbf{(Ours)} DistillH-Mamba-Student & \textbf{\textcolor{green}{\checkmark}} & \textbf{\textcolor{green}{\checkmark}} & 97.38 & \textbf{97.51} \\
\hline
\end{tabular}
\label{gsota}
\end{table*}

\subsection{Experiments Setting}

We implemented our Lightweight DistillH-Mamba architecture using PyTorch, with model training and evaluation conducted on a workstation equipped with an NVIDIA GeForce RTX 3080 Ti Laptop GPU (16GB VRAM), Intel Core i7-12700H CPU, and 32GB RAM. The implementation follows our two-stage approach: first training the full Hypergraph-Mamba teacher model, then using it to guide the training of the lightweight student model through relational knowledge distillation. For our evaluation protocol, we divided the 3D Skeletons UP-Fall dataset into three distinct subsets: a training set (80\%), validation set (10\%), and test set (10\%) for final performance assessment.

\section{Ablation Study}
Table \ref{ablation} presents a comprehensive ablation study comparing various model variants. The baseline Graph Convolutional Network (GCN) achieved 92.00\% accuracy with an inference time of 2.364×10$^{-1}$ seconds, while the Hypergraph Convolutional Network (HGCN) showed improved specificity (90.85\%) with slightly reduced inference time. To evaluate our dual-representation approach, we tested Hypergraph-Mamba variants with different processing pathways. The first-order processing (using matrix H) achieved 94.87\% accuracy, while second-order processing (using H$_2$) reached 95.13\%, indicating complementary information from both pathways. Combining both representations improved accuracy to 96.88\%.
Our final DistillH-Mamba student model, incorporating relational knowledge distillation, achieved the highest performance across all metrics: 97.38\% accuracy, 96.38\% precision, and 99.73\% specificity, while reducing inference time by 73.8\% compared to the teacher model. This demonstrates the effectiveness of knowledge distillation in both enhancing performance and improving computational efficiency.
\begin{table*}[ht]
\centering
\caption{Comparison with state-of-the-art methods on the UMAFall dataset.}
\begin{tabular}{lcccc}
\hline
\textbf{Method} & \textbf{Fall} & \textbf{Impact} & \textbf{Accuracy (\%)} & \textbf{F1-score (\%)} \\
\hline
LSTM\cite{mankodiya2022xai} & \textcolor{green}{\checkmark} & \text{\textcolor{red}{\sffamily X}} & 93.5 & - \\
FallNeXt\cite{mekruksavanich2022fallnext} & \textcolor{green}{\checkmark}& \text{\textcolor{red}{\sffamily X}} & 93.97 & 94.41 \\
\textbf{(Ours)} DistillH-Mamba-Student(Cross-Dataset Transfer Evaluation)  & \textbf{\textcolor{green}{\checkmark}} & \textbf{\textcolor{green}{\checkmark}} & \textbf{94.00} & \textbf{94.44} \\
\textbf{(Ours)} DistillH-Mamba-Student(Direct Training and Testing on UMAFall)  & \textbf{\textcolor{green}{\checkmark}} & \textbf{\textcolor{green}{\checkmark}} & \textbf{97.62} & \textbf{97.32} \\
\hline
\end{tabular}
\label{umafall}
\end{table*}
\subsection{Knowledge Distillation Effectiveness Analysis}

Our comprehensive ablation study (Table~\ref{tab:kd_ablation}) reveals the critical importance of knowledge distillation in achieving our final performance. Without knowledge distillation guidance, our student model achieves only 94.21\% accuracy with 89.45\% specificity, establishing a significant 2.67\% performance gap with the teacher model (96.88\% accuracy, 92.14\% specificity).
Our relational knowledge distillation framework substantially outperforms both baseline and standard approaches. While standard knowledge distillation achieves 95.67\% accuracy (+1.46\% over baseline), our relational KD reaches 97.38\% accuracy (+3.17\% over baseline), even surpassing teacher performance.
This apparent paradox where a smaller, faster student model outperforms its larger, more complex teacher is a well-documented phenomenon in knowledge distillation literature~\cite{park2019relational}, particularly in relational knowledge distillation frameworks that focus on preserving structural relationships rather than output mimicking.This enhancement of the student relies on three established theoretical factors: First, relational knowledge distillation acts as regularization, reducing overfitting present in the more complex teacher model. The teacher's higher parameter count (70.12M vs 23.34M) makes it more susceptible to memorizing training patterns rather than learning generalizable representations. Second, preserving critical spatial-temporal relationships rather than merely mimicking outputs focuses the student model on the most discriminative features.
Our relational losses (Equations 14-15) transfer structural information that cannot be encoded in conventional output matching, enabling the student to learn more refined impact detection patterns. Third, the reduced parameter space forces more efficient representations, eliminating redundant features that may have introduced noise. This architectural constraint, combined with our dual-pathway hypergraph design, creates a more specialized and efficient model for impact detection tasks.
Most significantly, our approach achieves specificity improvement from 89.45\% to 99.73\% (+10.28\%), critical for minimizing false alarms in emergency response systems. These results demonstrate that relational knowledge distillation transforms an underperforming compact model into one exceeding teacher capabilities while achieving substantial computational efficiency gains (73.8\% inference time reduction, 82.83\% FLOPs reduction), resolving the apparent performance-efficiency contradiction through theoretically grounded knowledge transfer mechanisms.

\subsection{Comparison with the State of the Art}

We evaluate DistillH-Mamba against state-of-the-art methods using two comprehensive comparisons to demonstrate both domain-specific and general competitiveness.

\subsubsection{Comparison with Skeleton-based Methods}
Table \ref{sota} presents a direct comparison with recent skeleton-based fall detection methods evaluated on the UP-Fall dataset. Our DistillH-Mamba approach achieves 97.38\% accuracy and 97.51\% F1-score, demonstrating strong performance within the skeleton-based domain. While our accuracy is slightly lower than CNNs \cite{galvao2021multimodal} (98.62\%), our superior F1-score (97.51\% vs 93.00\%) indicates better precision-recall balance, crucial for reliable fall detection systems. We significantly outperform other skeleton-based approaches including ST-GCN \cite{yan2023skeleton} (83.57\% F1-score) and BERT-based methods \cite{ramirez2023bert} (80.95\% F1-score).
Importantly, DistillH-Mamba is the only skeleton-based approach that explicitly detects impact events within falls, enabling identification of precisely when the body contacts the ground a capability crucial for emergency response systems that existing skeleton-based methods lack.

\subsubsection{Comparison with General Fall Detection Methods}
To position our work within the broader fall detection landscape, Table \ref{gsota} presents a comparison between our DistillH-Mamba with recent state-of-the-art fall detection methods across different datasets and modalities. While some approaches report higher accuracy on their respective datasets FL-FD \cite{qi2023fl} (99.92\%) and Fall-Mamba \cite{zhang2025fall} (99.63\%) these results are not directly comparable due to different datasets and evaluation protocols. Our approach demonstrates competitive performance (97.38\% accuracy, 97.51\% F1-score) while providing unique impact detection capabilities that none of these general fall detection methods offer. This cross-modal comparison highlights that DistillH-Mamba achieves strong performance using only skeletal data, whereas many competing approaches rely on multimodal inputs or different data types.

\subsection{Cross-Dataset Validation on UMAFall Dataset}

To validate the generalizability and robustness of our proposed DistillH-Mamba model,we conducted two comprehensive evaluations using the UMAFall dataset.
\begin{figure*}[t]
\centering
\includegraphics[width=2\columnwidth]{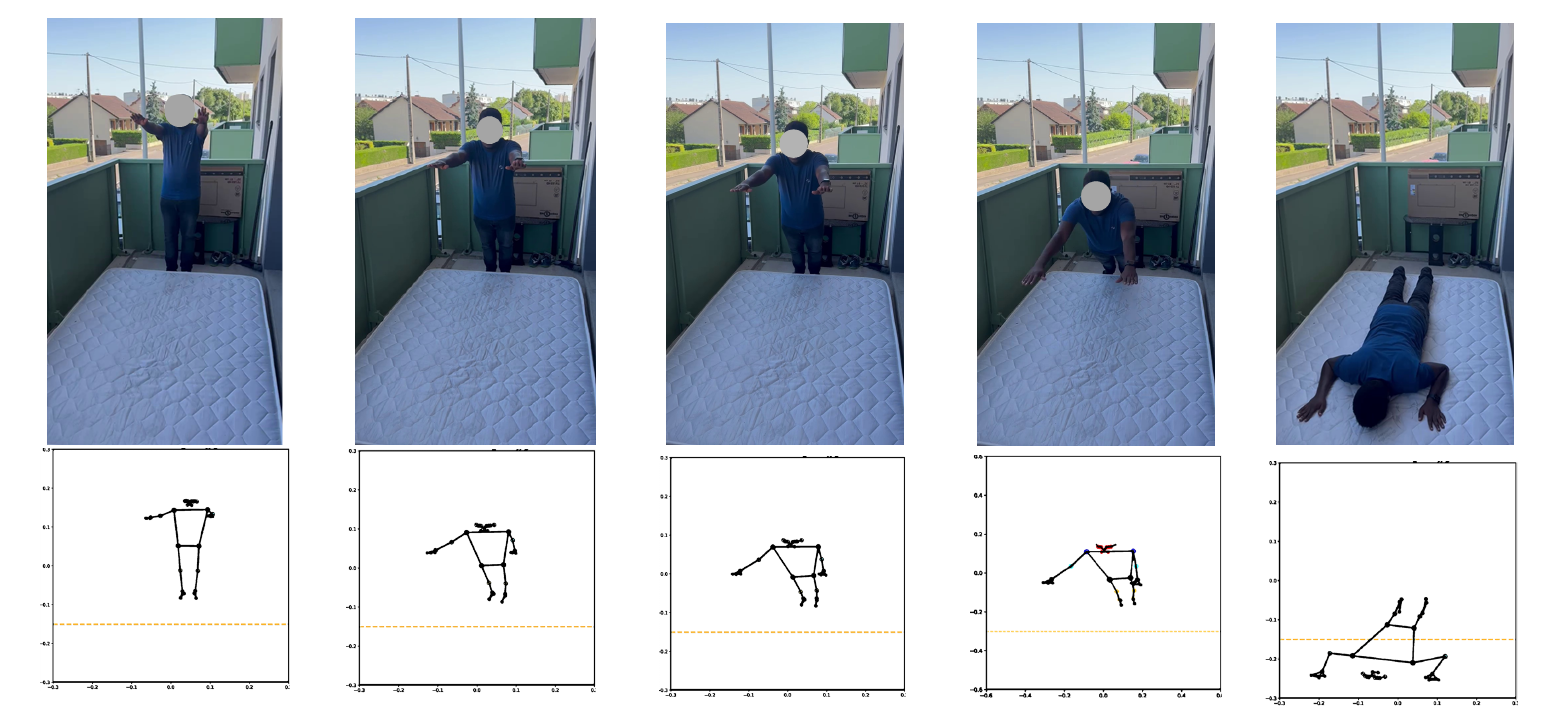}
\caption{Representative frames from self-created validation dataset showing impact fall progression: (top row) original frames capturing the fall sequence from standing to ground contact, (bottom row) corresponding extracted skeletal representations used for impact detection evaluation.}
\label{self_created}
\end{figure*}
\subsubsection{Cross-Dataset Transfer Evaluation}

To evaluate generalizability, we conducted cross-dataset testing by training our model on the 3D Skeletons UP-Fall dataset and applying it directly to the UMAFall test set without fine-tuning a rigorous test of knowledge transfer across varying environments.
Despite inherent differences between datasets (camera angles, lighting, subject behavior), our model achieved 94.00\% accuracy and 94.44\% F1-score with high specificity (93.22\%), crucial for minimizing false alarms in real-world monitoring.
As shown in Table \ref{umafall}, our method outperforms state-of-the-art models trained directly on UMAFall, including Mekruksavanich et al. \cite{mekruksavanich2022fallnext} (93.97\% accuracy, 94.41\% F1) and Mankodiya et al. \cite{mankodiya2022xai} (93.5\% accuracy).
These results highlight the strength of our dual-representation hypergraph architecture and Mamba-based modeling, demonstrating superior cross-dataset generalization for precise impact detection across diverse settings.

\begin{table}[t]
\centering
\caption{Performance Comparison: Benchmark vs. Self-Created Dataset}
\label{self_created_comparison}
\begin{tabular}{lccc}
\toprule
\textbf{Metric} & \textbf{UMAFall} & \textbf{UP-Fall} & \textbf{Self-Created} \\
\midrule
Accuracy (\%) & 97.62 & 97.38 & 91.33 \\
Precision (\%) & - & 96.38 & 89.00 \\
Recall (\%) & 97.32 & 98.84 & 94.89 \\
Specificity (\%) & - & 99.73 & 88.20 \\
F1-Score (\%) & 97.32 & 97.51 & 91.85 \\
\bottomrule
\end{tabular}
\end{table}

\subsubsection{Direct Training and Testing on UMAFall}

When directly trained and tested on the UMAFall dataset, our DistillH-Mamba-Student model achieves even higher performance of 97.62\% accuracy and 97.32\% F1 score as shown in Table \ref{umafall}.
This performance significantly exceeds both our cross-dataset transfer results (94.00\% accuracy) and state-of-the-art methods specifically designed for UMAFall, including \cite{mekruksavanich2022fallnext} (93.97\% accuracy) and \cite{mankodiya2022xai} (93.5\% accuracy). These results further validate our approach's effectiveness across different datasets and experimental conditions.

\subsection{Self-Created Dataset Validation}

To evaluate model generalization beyond standard benchmarks, we constructed a proprietary validation set by recording and annotating real human motion, including simulated fall events. Specifically, one of the authors performed controlled fall simulations in an indoor environment, captured using an RGB camera at 30 fps. The recorded sequence comprises 300 frames, structured into 3 temporal batches of 100 frames each, with skeletal data represented as 33 joints in 3D space (x, y, z) coordinates, consistent with the model's input format. The pre-trained DistillH-Mamba-Student model was evaluated in zero-shot mode without fine-tuning.
As illustrated in Figure \ref{self_created}, the validation dataset captures realistic fall dynamics from standing posture through ground contact. Table \ref{self_created_comparison} presents the performance comparison across datasets. The model achieved 91.33\% accuracy on this self-created dataset, demonstrating good generalization capability beyond benchmark datasets. The performance shows 89.00\% precision and 94.89\% recall, with an F1-score of 91.85\%. The 88.20\% specificity indicates reasonable ability to correctly identify non-impact events. While the performance decrease compared to benchmark datasets is expected due to the more challenging and uncontrolled nature of self-created data, these results validate the model's robustness on independent data while maintaining competitive performance across different evaluation conditions.

\subsection{Visual Analysis of Knowledge Transfer}

To validate our knowledge distillation approach, we visualize and compare attention patterns and feature representations between teacher and student models. This analysis reveals how spatial relationships and temporal dynamics are effectively preserved despite the significant reduction in model complexity.

\subsubsection{Comparative Analysis of Spatial Attention Patterns}

Figure \ref{spatial} presents a comparative analysis of spatial attention patterns between teacher and student models. In first-order processing ($\mathbf{OH}$ features), the teacher model exhibits varying activation strengths, with joints 0 (nose), 12 (left shoulder), and 31 (right foot index) showing strongest activations. These joints are most responsible for impact during falls the nose being a primary impact point in forward falls, the left shoulder indicating upper body positioning, and the right foot index revealing lower extremity dynamics during fall trajectory. The student model demonstrates similar patterns with a correlation coefficient of 0.8090, though with reduced intensity.
The second-order processing visualizations ($\mathbf{OH}_2$ features) highlight critical joint interactions with particularly strong activations at joints 3 (right eye outer), 6 (left eye outer), and 9 (mouth right). These facial feature joints are essential for detecting head positioning during falls, where monitoring head trajectory is crucial for assessing fall severity. Most notably, the student model maintains a high correlation coefficient of 0.9947 with the teacher model in second-order processing. This effective knowledge transfer demonstrates our approach's efficiency in maintaining detection performance while significantly reducing computational demands.

\begin{center}
\includegraphics[width=8cm]{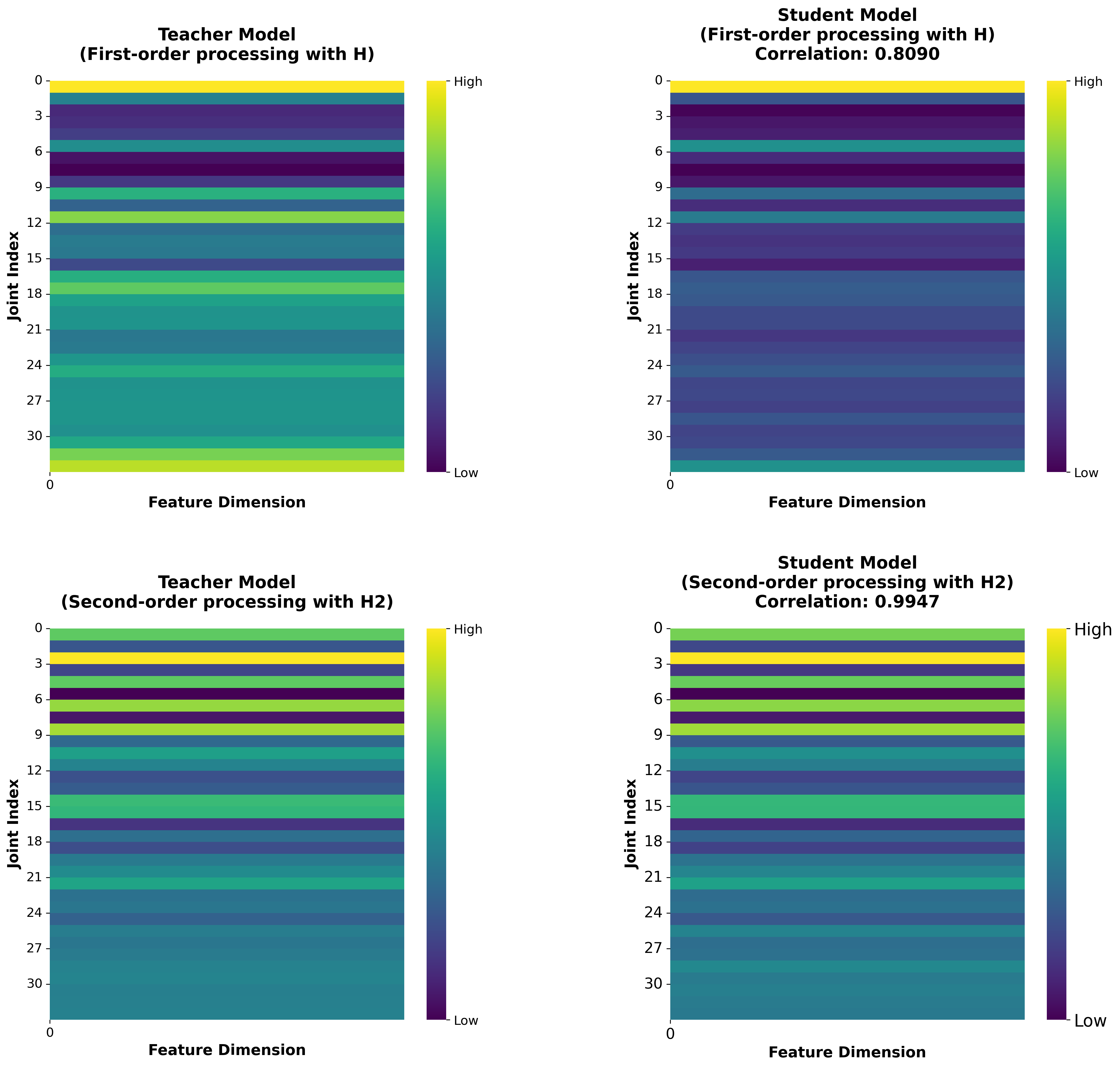}
\captionof{figure}{Spatial attention visualization of $\mathbf{OH}$ and $\mathbf{OH}_2$ features in teacher and student models}
\label{spatial}
\end{center}

\begin{center}
\begin{table*}
\caption{Computational Efficiency Comparison Between DistillH-Mamba Models and STGCN using 3D Skeletons UP-Fall Dataset}
\label{efficiency_comparison}
\begin{tabular}{lcccc}
\toprule
\textbf{Metric} & \textbf{DistillH-Mamba-Teacher} & \textbf{DistillH-Mamba-Student} & \textbf{Improvement}$^{\ddagger}$ (\%) & \textbf{STGCN \cite{keskes2021vision}} \\
\midrule
Parameters (M) & 70.115729 & 23.341701 & 66.00 & 107.858705 \\
Memory Footprint (MB) & 280.46 & 93.37 & 66.70 & 431.43 \\
FLOPs & $2.266 \times 10^7$ & $3.891 \times 10^6$ & 82.83 & $9.717 \times 10^7$ \\
Inference Time (s) & $1.823 \times 10^{-1}$ & $0.478 \times 10^{-1}$ & 73.80$^{*}$ & $6.208 \times 10^{-1}$ \\
Inference Speed (samples/s) & $2.332 \times 10^3$ & $4.592 \times 10^3$ & 97.02$^{\dagger}$ & $12.618 \times 10^3$ \\
\bottomrule
\end{tabular}
\vspace{2mm}

\footnotesize
$^{*}$Time reduction: lower is better \quad
$^{\dagger}$Speed improvement: higher is better \quad
$^{\ddagger}$Improvement between DistillH-Mamba-Teacher and DistillH-Mamba-Student
\end{table*}
\end{center}

\subsubsection{Comparative Analysis of Temporal Features}

Figure \ref{temporal} visualizes temporal feature representations of joint 0 (nose) and joint 3 (right eye outer) over 100 time steps, demonstrating effective knowledge transfer between teacher and student models. These joints were selected based on their high activation levels in spatial attention patterns, allowing us to confirm whether the student model preserves critical spatio-temporal features during impact events.
The "First-order Processing with H" ($\mathbf{OH}$ features) shows exceptional temporal knowledge transfer, with both joints achieving perfect correlation coefficient ($r = 1.000$) and low root mean square error rates (RMSE = 0.050 for joint 0 and RMSE = 0.045 for joint 3). The "Second-order Processing with H2" ($\mathbf{OH}_2$ features) also demonstrates perfect correlation for joint 0 ($r = 1.000$, RMSE = 0.050) and joint 3 ($r = 1.000$, RMSE = 0.054).
The visualization clearly shows that despite significant model compression, the student model precisely maintains the temporal dynamics of these critical joints throughout the fall sequence. The purple shaded areas represent the range of potential feature values, while the nearly identical teacher (blue) and student (red) feature trajectories confirm exceptional preservation of temporal patterns.
The two processing approaches reveal complementary aspects of temporal impact dynamics: first-order processing captures direct joint movements while second-order processing reveals more complex oscillatory patterns characteristic of impact events. These results demonstrate that our relational distillation approach effectively preserves critical spatio-temporal dynamics necessary for accurate impact detection while significantly reducing computational requirements.

\begin{center}
\includegraphics[width=8.5cm]{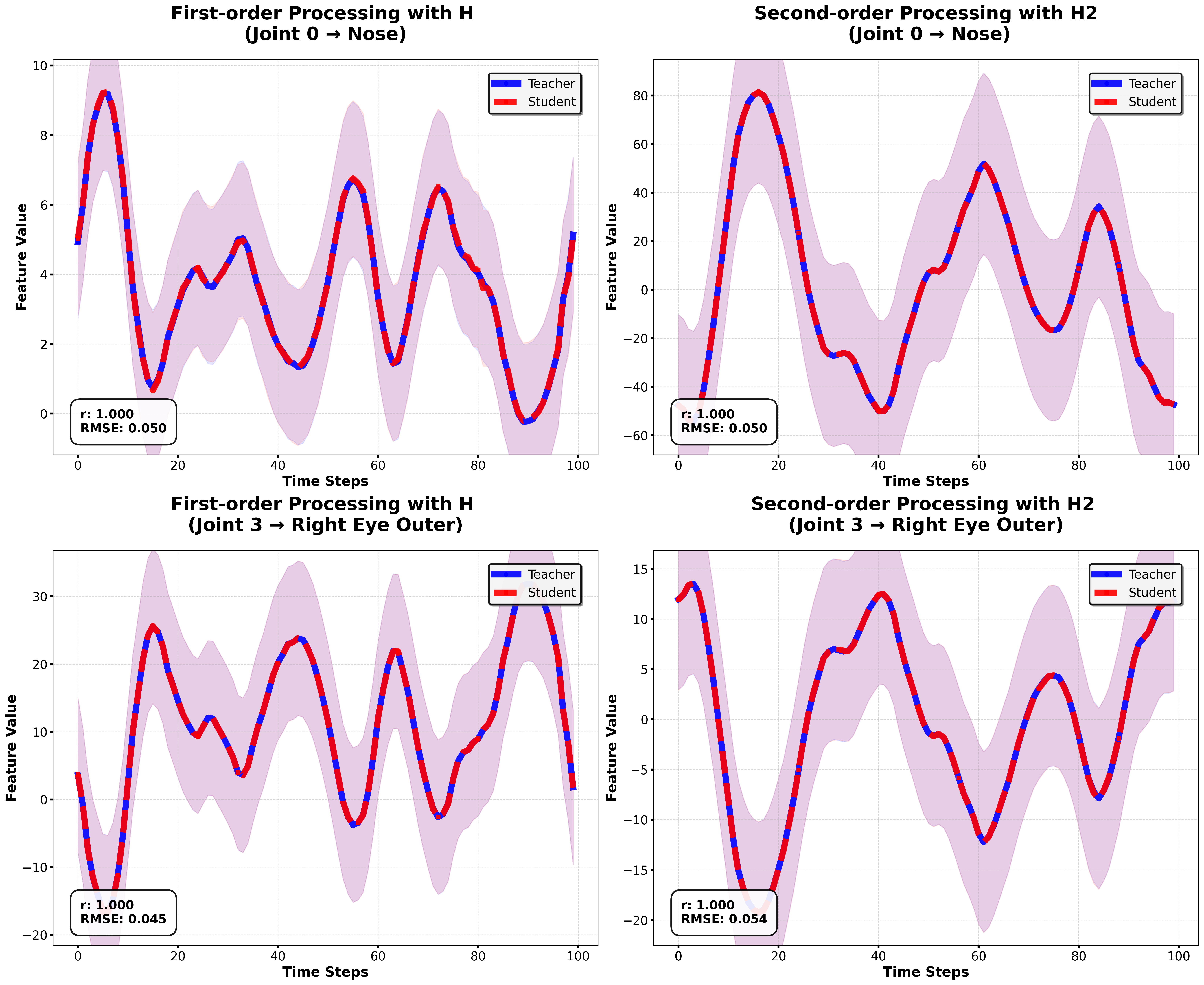}
\captionof{figure}{Temporal feature visualization of $\mathbf{OH}$ and $\mathbf{OH}_2$ features for joints 0 (Nose) and 3 (Right eye outer) in teacher (blue) and student (red) models, with correlation coefficients (r) and RMSE values}
\label{temporal}
\end{center}

\subsection{Computational Efficiency Analysis}
% \color{black}  % or \normalcolor

A primary contribution of our work is developing a lightweight impact detection model with significantly reduced computational requirements while maintaining high detection accuracy. Table \ref{efficiency_comparison} compares computational efficiency metrics between our proposed models and STGCN \cite{keskes2021vision}, a state-of-the-art baseline architecture for skeleton-based fall detection that we reimplemented following the original design for comparison.
The DistillH-Mamba-Student model achieves substantial efficiency improvements across all metrics compared to the teacher model. The parameter count is reduced by 66\%, decreasing from 70.12M to 23.34M parameters, with a corresponding memory footprint reduction from 280.46 MB to 93.37 MB (66.70\% reduction). This parameter reduction directly contributes to an 82.83\% decrease in floating-point operations (FLOPs). From a practical deployment perspective, these improvements translate to a 73.80\% reduction in inference time per sequence, enabling the student model to process nearly twice as many samples per second (97.02\% improvement in inference speed).
When compared to STGCN, our student model demonstrates even more dramatic efficiency gains. The memory footprint is 4.6× smaller (93.37 MB vs 431.43 MB), with significantly fewer parameters than STGCN's 107.86M and approximately 12 times faster inference speed. The 93.37 MB memory requirement is well within the capabilities of smartphones (4-12GB RAM), Raspberry Pi devices (4-8GB RAM), and dedicated healthcare monitors (1-2GB RAM) \cite{liebherr2020smartphones, rzepka2024performance}. This processing efficiency, combined with the reduced memory footprint, makes our lightweight model suitable for deployment on resource-constrained devices such as wearable sensors or edge computing platforms a critical requirement for real-time fall monitoring systems in healthcare applications.

\section{Conclusion}
\label{sec:Conclusion}
This study proposed DistillH-Mamba, a novel architecture for impact fall detection that identifies the precise moment when an individual contacts the ground during a fall by combining hypergraph-based spatial modeling with Mamba state-space temporal modeling, enhanced by relational knowledge distillation.
Our key contributions are: (1) a dual-representation hypergraph approach capturing complex higher-order relationships between skeletal joints; (2) integration of Mamba architecture with hypergraphs for efficient spatial-temporal processing; and (3) a relational knowledge distillation framework that preserves crucial patterns while reducing computational requirements.
Experimental results demonstrate good performance of our method over existing methods, with the lightweight student model surpassing its teacher across all metrics while achieving substantial computational efficiency gains. Cross-dataset evaluation confirmed its generalizability across environments.
By enabling precise impact detection with minimal resources, our approach facilitates deployment on resource-constrained devices for elderly care, supporting timely intervention. 
While our current validation demonstrates strong performance on publicly available datasets with simulated falls, we acknowledge the importance of real-world validation for clinical deployment. Future work will address current limitations by developing robust skeleton extraction methods, extending to multi-person detection, and conducting comprehensive validation studies with real elderly populations in naturalistic environments. We are actively pursuing collaborations with healthcare institutions and assisted living facilities to ethically collect and validate our approach on authentic fall data from elderly individuals, ensuring our system's effectiveness translates to real-world clinical settings where timely intervention is critical.
% Main content of your paper ends here
\bibliographystyle{IEEEtran}
\bibliography{reference}
% \bibliographystyle{unsrt}
% \bibliography{reference}

\end{document}